\pdfoutput=1

\documentclass[11pt]{article}

\usepackage{acl}

\usepackage{times}
\usepackage{latexsym}

\usepackage[T1]{fontenc}

\usepackage[utf8]{inputenc}

\usepackage{microtype}

\usepackage{amsmath}
\usepackage{amssymb}
\usepackage{booktabs}
\usepackage{fontawesome5}
\usepackage{graphicx}
\usepackage{multirow}
\usepackage{subcaption}
\usepackage{pdfcomment}

\newcommand{\bap}{\textsc{BAP}}

\newcommand{\depprobe}{\textsc{DepProbe}}
\newcommand{\nlpnorth}{\hspace{.2em}\faIcon{compass}}
\newcommand{\cis}{\resizebox{!}{.75em}{\faIcon{mountain}}}

\title{{\normalsize\faIcon{sort-amount-down}} Sort by Structure: Language Model Ranking as Dependency Probing}

\author{
    Max M{\"u}ller-Eberstein\textsuperscript{\nlpnorth{}}  \and
    Rob van der Goot\textsuperscript{\nlpnorth{}} \and
    Barbara Plank\textsuperscript{\nlpnorth{}\cis{}} \\
    \textsuperscript{\nlpnorth{}} Department of Computer Science, IT University of Copenhagen, Denmark \\
    \textsuperscript{\cis{}} Center for Information and Language Processing (CIS), LMU Munich, Germany \\
    \texttt{mamy@itu.dk, robv@itu.dk, bplank@cis.lmu.de} \\}

\begin{document}
\maketitle
\begin{abstract}
Making an informed choice of pre-trained language model (LM) is critical for performance, yet environmentally costly, and as such widely underexplored. The field of Computer Vision has begun to tackle encoder ranking, with promising forays into Natural Language Processing, however they lack coverage of linguistic tasks such as structured prediction. We propose \emph{probing to rank LMs}, specifically for parsing dependencies in a given language, by measuring the degree to which labeled trees are recoverable from an LM's contextualized embeddings. Across 46 typologically and architecturally diverse LM-language pairs, our probing approach predicts the best LM choice 79\% of the time using orders of magnitude less compute than training a full parser. Within this study, we identify and analyze one recently proposed decoupled LM---RemBERT---and find it strikingly contains less inherent dependency information, but often yields the best parser after full fine-tuning. Without this outlier our approach identifies the best LM in 89\% of cases.
\end{abstract}

\section{Introduction}

With the advent of massively pre-trained language models (LMs) in Natural Language Processing (NLP), it has become crucial for practitioners to choose the best LM encoder for their given task early on, regardless of the rest of their proposed model architecture. The greatest variation of LMs lies in the language or domain-specificity of the unlabelled data used during pre-training (with architectures often staying identical).

Typically, better expressivity is expected from language/domain-specific LMs \citep{gururangan-etal-2020-dont,dai-etal-2020-cost} while open-domain settings necessitate high-capacity models with access to as much pre-training data as possible. This tradeoff is difficult to navigate, and given that multiple specialized LMs (or none at all) are available, practitioners often resort to an ad-hoc choice. In absence of immediate performance indicators, the most accurate choice could be made by training the full model using each LM candidate, however this is often infeasible and wasteful~\cite{strubell-etal-2019-energy}.

Recently, the field of Computer Vision (CV) has attempted to tackle this problem by quantifying useful information in pre-trained image encoders as measured directly on labeled target data without fine-tuning \citep{nguyen2020,you2021}. While first forays for applying these methods to NLP are promising, some linguistic tasks differ substantially: Structured prediction, such as parsing syntactic dependencies, is a fundamental NLP task not covered by prior encoder ranking methods due to its graphical output. Simultaneously, performance prediction in NLP has so far been studied as a function of dataset and model characteristics~\citep{xia-etal-2020-predicting,ye-etal-2021-towards} and has yet to examine how to rank large pools of pre-trained LMs.

Given the closely related field of probing, in which lightweight models quantify task-specific information in pre-trained LMs, we recast its objective in the context of performance prediction and ask: \textit{How predictive is lightweight probing at choosing the best performing LM for dependency parsing?} To answer this question, we contribute:

\begin{itemize}
    \item An efficient encoder ranking method for structured prediction using dependency probing (\citealp{depprobe}; \depprobe{}) to quantify latent syntax (Section \ref{sec:method}).
    \item Experiments across 46 typologically and architecturally diverse LM + target language combinations (Section \ref{sec:experiments}).\footnote{Code at \href{https://personads.me/x/naacl-2022-code}{https://personads.me/x/naacl-2022-code}.}
    \item An in-depth analysis of the surprisingly low inherent dependency information in RemBERT ~\citep{chung2021rethinking} compared to its high fine-tuned performance (Section \ref{sec:rembert-analysis}).
\end{itemize}

\section{Methodology}\label{sec:method}

Probing pre-trained LMs is highly related to encoder ranking in CV where the ease of recoverability of class-differentiating information is key \citep{nguyen2020,you2021}. This approach is more immediate than existing NLP performance prediction methods which rely on featurized representations of source and target data without actively ranking encoders \citep{xia-etal-2020-predicting,ye-etal-2021-towards}. As most experiments in NLP are conducted using a limited set of LMs---often a single model---without strong prior motivations, \textit{we see LM ranking as a critical task on its own.}

While probes for LMs come in many forms, they are generally characterized as lightweight, minimal architectures intended to solve a particular task \citep{hall-maudslay-etal-2020-tale}. While non-linear models such as small multi-layer perceptrons are often used \citep{tenney2018what}, there have been criticisms given that their performance highly depends on the complexity of their architecture \citep{hewitt-liang-2019-designing,voita-titov-2020-information}. As such, we rely on linear probes alone, which have the benefit of being extremely lightweight, closely resembling existing performance prediction methods \citep{you2021}, and allow for statements about linear subspaces contained in LM latent spaces.

\begin{figure}
    \centering
    \pdftooltip{\includegraphics[width=.47\textwidth]{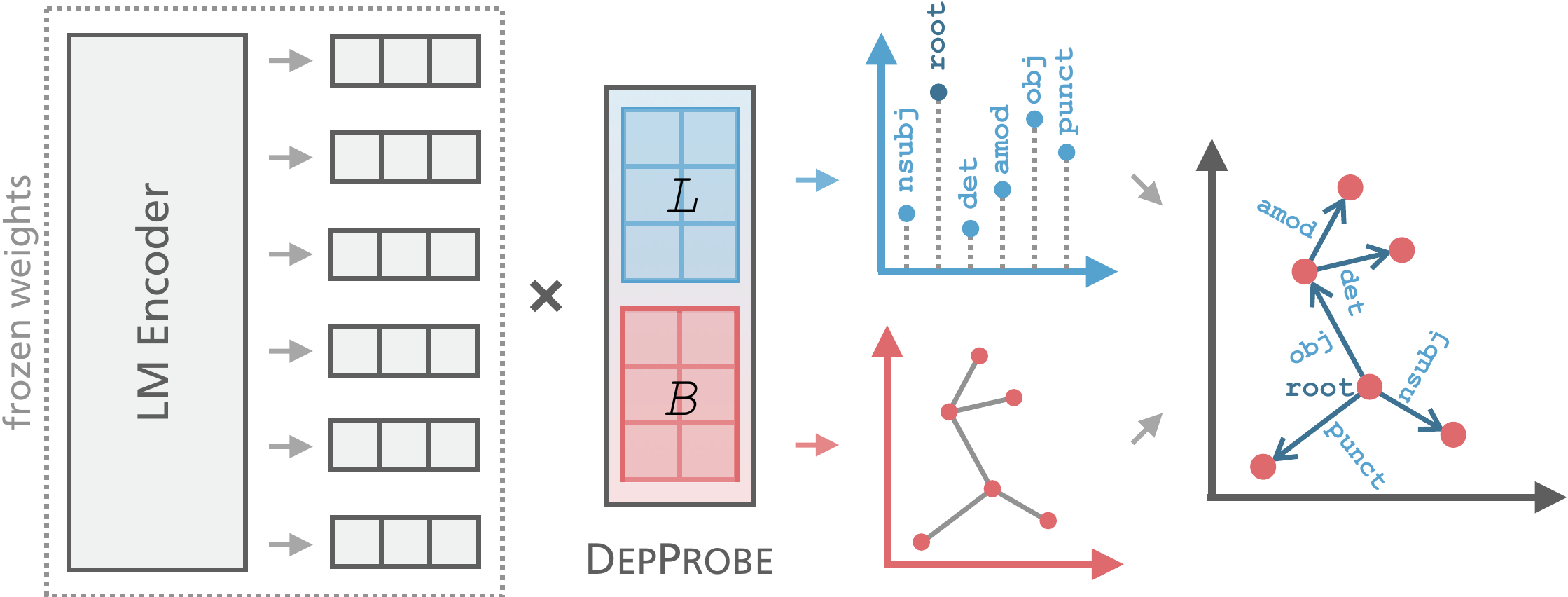}}{Screenreader Caption: Frozen embeddings are multiplied with matrices B and L to transform them into a tree structural subspace and label subspace. Undirected graph extracted from MST in structural space. Node with highest root probability in label space sets directionality of all edges by pointing them away. Edges are labeled according to child embedding label from L.}
    \caption{\textbf{Visualization of \depprobe{}.} Relational and structural subspaces $L$ and $B$ are combined to extract labeled, directed trees from embeddings.}
    \label{fig:depprobe}
\end{figure}

\paragraph{\depprobe{}} (\citealp{depprobe}; visualized in Figure \ref{fig:depprobe}) is a linear formulation for extracting fully labeled dependency trees based on the structural probe by \citet{hewitt-manning-2019-structural}. Given contextualized embeddings of dimensionality $d$, a linear transformation $B \in \mathbb{R}^{b \times d}$ with $b \ll d$ (typically $b$ = 128) maps them into a subspace in which the Euclidean distance between embeddings corresponds to the number of edges between the respective words in the gold dependency graph.

In our formulation, we supplement a linear transformation $L \in \mathbb{R}^{l \times d}$ (with $l$ = number of dependency relations) which maps each embedding to a subspace in which the magnitude of each dimension corresponds to the likelihood of a word and its head being governed by a certain relation.

By computing the minimum spanning tree in $B$ and then finding the word with the highest root likelihood in $L$, we can determine the directionality of all edges as pointing away from the root. All remaining edges are labeled according to the most likely non-root class in $L$, resulting in a fully directed and labeled dependency tree.

Note that this approach differs substantially from prior approaches which yield undirected and/or unlabeled trees \citep{hewitt-manning-2019-structural,kulmizev-etal-2020-neural} or use pre-computed edges and non-linear classifiers \citep{tenney2018what}. \depprobe{} efficiently computes the full target metric (i.e.\ labeled attachment scores) instead of approximate alternatives (e.g.\ undirected, unlabeled attachment scores or tree depth correlation).

\section{Experiments}\label{sec:experiments}

\paragraph{Setup} We investigate the ability of \depprobe{} to select the best performing LM for dependency parsing across nine linguistically diverse treebanks from Universal Dependencies (\citealp{ud29}; UD) which were previously chosen by \citet{smith-etal-2018-investigation} to reflect diverse writing systems and morphological complexity (see Appendix \ref{sec:treebank-details}).

For each target language, we employ three multilingual LMs---mBERT \citep{devlin-etal-2019-bert}, XLM-R \citep{conneau-etal-2020-unsupervised}, RemBERT \citep{chung2021rethinking}---as well as 1--3 language-specific LMs retrieved by popularity from HuggingFace's Model Hub \citep{wolf-etal-2020-transformers}, resulting in a total of 46 LM-target pair setups (see Appendix \ref{sec:result-details}).

For each combination, we train a \depprobe{} to compute labeled attachment scores (LAS), hypothesizing that LMs from which trees are most accurately recoverable also perform better in a fully tuned parser. To evaluate the true downstream performance of a fully-tuned model, we further train a deep biaffine attention parser (\bap{};   \citealp{dozat2017}) on each LM-target combination. Compared to full fine-tuning, \depprobe{} only optimizes the matrices $B$ and $L$, resulting in the extraction of labeled trees with as few as 190k instead of 583M trainable parameters for the largest RemBERT model (details in Appendix \ref{sec:experiment-details}).

We measure the predictive power of probing for fully fine-tuned model performance using the Pearson correlation coefficient $\rho$ as well as the weighted Kendall's $\tau_w$ \citep{vigna2015}. The latter metric corresponds to a correlation coefficient in $[-1,1]$ and simultaneously defines the probability of choosing the better LM given a pair as $\frac{\tau_w + 1}{2}$, allowing us to quantify the overall quality of a ranking.

\begin{figure}
    \centering
    \begin{subfigure}[c]{.37\textwidth}
        \hspace{-.7em}
        \pdftooltip{\includegraphics[width=\textwidth]{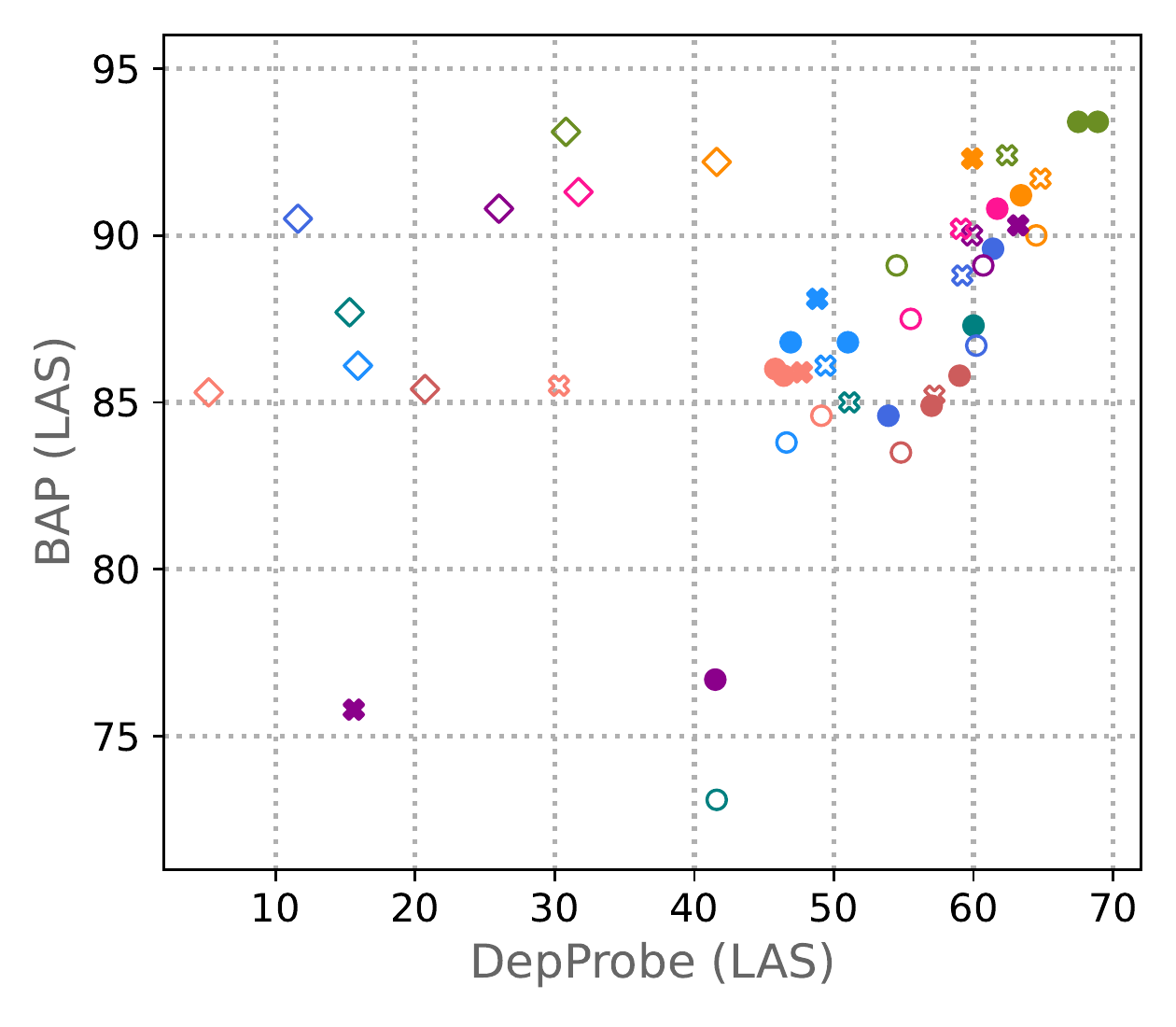}}{Screenreader Caption: BAP and DepProbe (LAS). bert-base-multilingual-cased on AR-PADT: 83.50 and 54.80, xlm-roberta-base on AR-PADT: 85.20 and 57.20, google/rembert on AR-PADT: 85.40 and 20.70, aubmindlab/bert-base-arabertv02 on AR-PADT: 85.80 and 59.00, asafaya/bert-base-arabic on AR-PADT: 84.90 and 57.00, bert-base-multilingual-cased on EN-EWT: 90.00 and 64.50, xlm-roberta-base on EN-EWT: 91.70 and 64.80, google/rembert on EN-EWT: 92.20 and 41.60, bert-base-uncased on EN-EWT: 91.20 and 63.40, roberta-large on EN-EWT: 92.30 and 59.90, bert-base-multilingual-cased on FI-TDT: 89.10 and 54.50, xlm-roberta-base on FI-TDT: 92.40 and 62.40, google/rembert on FI-TDT: 93.10 and 30.80, TurkuNLP/bert-base-finnish-uncased-v1 on FI-TDT: 93.40 and 68.90, TurkuNLP/bert-base-finnish-cased-v1 on FI-TDT: 93.40 and 67.50, bert-base-multilingual-cased on GRC-PROIEL: 73.10 and 41.60, xlm-roberta-base on GRC-PROIEL: 85.00 and 51.10, google/rembert on GRC-PROIEL: 87.70 and 15.30, pranaydeeps/Ancient-Greek-BERT on GRC-PROIEL: 87.30 and 60.00, nlpaueb/bert-base-greek-uncased-v1 on HE-HTB: 84.60 and 53.90, bert-base-multilingual-cased on HE-HTB: 86.70 and 60.20, xlm-roberta-base on HE-HTB: 88.80 and 59.20, google/rembert on HE-HTB: 90.50 and 11.60, onlplab/alephbert-base on HE-HTB: 89.60 and 61.40, bert-base-multilingual-cased on KO-GSD: 83.80 and 46.60, xlm-roberta-base on KO-GSD: 86.10 and 49.40, google/rembert on KO-GSD: 86.10 and 15.90, klue/bert-base on KO-GSD: 86.80 and 51.00, klue/roberta-large on KO-GSD: 88.10 and 48.80, kykim/bert-kor-base on KO-GSD: 86.80 and 46.90, bert-base-multilingual-cased on RU-GSD: 89.10 and 60.70, xlm-roberta-base on RU-GSD: 90.00 and 59.90, google/rembert on RU-GSD: 90.80 and 26.00, cointegrated/rubert-tiny on RU-GSD: 76.70 and 41.50, sberbank-ai/ruRoberta-large on RU-GSD: 90.30 and 63.20, blinoff/roberta-base-russian-v0 on RU-GSD: 75.80 and 15.60, bert-base-multilingual-cased on SV-Talbanken: 87.50 and 55.50, xlm-roberta-base on SV-Talbanken: 90.20 and 59.10, google/rembert on SV-Talbanken: 91.30 and 31.70, KB/bert-base-swedish-cased on SV-Talbanken: 90.80 and 61.70, bert-base-multilingual-cased on ZH-GSD: 84.60 and 49.10, xlm-roberta-base on ZH-GSD: 85.50 and 30.30, google/rembert on ZH-GSD: 85.30 and 5.20, bert-base-chinese on ZH-GSD: 85.80 and 46.40, hfl/chinese-bert-wwm-ext on ZH-GSD: 86.00 and 45.80, hfl/chinese-roberta-wwm-ext on ZH-GSD: 85.90 and 47.70.}
    \end{subfigure}
    \hspace{-1em}
    \begin{subfigure}[c]{.08\textwidth}
        \includegraphics[width=\textwidth]{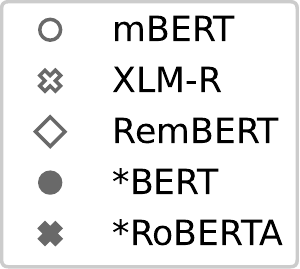}\\[.8em]
        \includegraphics[width=.7\textwidth]{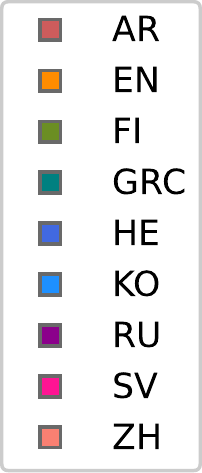}
    \end{subfigure}
    \vspace{-.5em}
    \caption{\textbf{LAS of \depprobe{} in relation to full \bap{}} across nine language targets (dev) using language-specific and multilingual LM encoders of different architecture types (exact scores in Appendix \ref{sec:result-details}).}
    \label{fig:scatter-results}
\end{figure}

\paragraph{Results}

Comparing the LAS of \depprobe{}'s lightweight predictions against full \bap{} fine-tuning in Figure \ref{fig:scatter-results}, we see a clear correlation as the probe correctly predicts the difficulty of parsing languages relative to each other and also ranks models within languages closely according to their final performance. With a $\tau_w$ of .58 between scores ($p < 0.001$), this works out to \depprobe{} selecting the better performing final model given any two models 79\% of the time. Additionally, LAS is slightly more predictive of final performance than unlabeled, undirected attachment scores (UUAS) with $\tau_w$ = .57 to which prior probing approaches are restricted (see Appendix \ref{sec:result-details}).

Given a modest $\rho$ of .32 ($p < 0.05$), we surprisingly also observe a single strong outlier to this pattern, namely the multilingual RemBERT \citep{chung2021rethinking} decoupled LM architecture. While \depprobe{} consistently ranks it low as it cannot extract dependency parse trees as accurately as from the BERT and RoBERTa-based architectures, RemBERT actually performs best on four out of the nine targets when fully fine-tuned in \bap{}. Excluding monolingual LMs, it further outperforms the other multilingual LMs in seven out of nine cases. As it is a more recent and distinctive architecture with many differences to the most commonly-used contemporary LMs, we analyze potential reasons for this discrepancy in Section \ref{sec:rembert-analysis}.

Excluding RemBERT as an outlier, we find substantially higher correlation among all other models: $\rho$ = .78 and $\tau_w$ = .78 ($p< 0.001$). This means that among these models, fully fine-tuning the LM for which \depprobe{} extracts the highest scores, yields the better final performance 89\% of the time.

In practice, learning \depprobe{}'s linear transformations while keeping the LM frozen is multiple orders of magnitude more efficient than fully training a complex parser plus the LM's parameters. As such, linear probing offers a viable method for selecting the best encoder in absence of qualitative heuristics or intuitions. This predictive performance is furthermore achievable in minutes compared to hours and at a far lower energy budget (see Appendices \ref{sec:experiment-details} and \ref{sec:result-details}).

\section{Probing
Decoupled LMs}\label{sec:rembert-analysis}

Considering \depprobe{}'s high predictive performance across LMs with varying architecture types, languages/domains and pre-training procedures, we next investigate its limitations: Specifically, which differences in RemBERT \citep{chung2021rethinking} lead to it being measured as an outlier with seemingly low amounts of latent dependency information despite reaching some of the highest scores after full fine-tuning. The architecture has 32 layers and embeddings with $d$ = 1152, compared to most models' 12 layers and $d$ = 768. It accommodates these size and depth increases within a manageable parameter envelope by using smaller input embeddings with $d_{\text{in}}$ = 256. While choosing different $d$ for the input and output embeddings is not possible in most prior models due to both embedding matrices being coupled, RemBERT decouples them, leading to a larger parameter budget and less overfitting on the masked language modeling pre-training task \citep{chung2021rethinking}.

\begin{figure}
    \centering
\pdftooltip{\includegraphics[width=.47\textwidth]{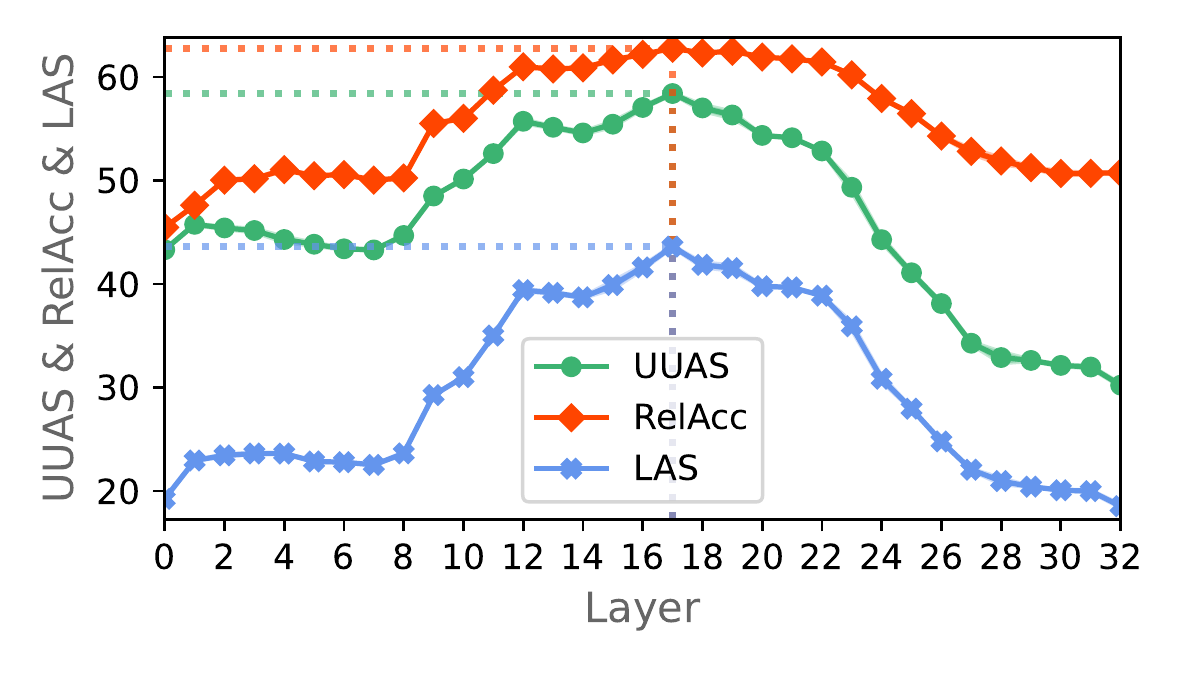}}{Screenreader Caption: UUAS: 43.3 at l=0, 45.8 at l=1, 45.4 at l=2, 45.2 at l=3, 44.3 at l=4, 43.8 at l=5, 43.4 at l=6, 43.3 at l=7, 44.7 at l=8, 48.5 at l=9, 50.1 at l=10, 52.6 at l=11, 55.7 at l=12, 55.1 at l=13, 54.6 at l=14, 55.4 at l=15, 57.1 at l=16, 58.4 at l=17, 57.0 at l=18, 56.3 at l=19, 54.4 at l=20, 54.1 at l=21, 52.9 at l=22, 49.3 at l=23, 44.3 at l=24, 41.1 at l=25, 38.1 at l=26, 34.3 at l=27, 32.9 at l=28, 32.6 at l=29, 32.1 at l=30, 32.0 at l=31, 30.2 at l=32. RelAcc: 45.5 at l=0, 47.6 at l=1, 50.0 at l=2, 50.2 at l=3, 51.0 at l=4, 50.5 at l=5, 50.6 at l=6, 50.0 at l=7, 50.2 at l=8, 55.5 at l=9, 56.0 at l=10, 58.7 at l=11, 61.0 at l=12, 60.8 at l=13, 60.9 at l=14, 61.6 at l=15, 62.2 at l=16, 62.8 at l=17, 62.3 at l=18, 62.5 at l=19, 61.9 at l=20, 61.8 at l=21, 61.5 at l=22, 60.2 at l=23, 57.9 at l=24, 56.5 at l=25, 54.3 at l=26, 52.8 at l=27, 51.9 at l=28, 51.2 at l=29, 50.7 at l=30, 50.7 at l=31, 50.8 at l=32. LAS: 19.2 at l=0, 22.9 at l=1, 23.4 at l=2, 23.6 at l=3, 23.6 at l=4, 22.9 at l=5, 22.8 at l=6, 22.5 at l=7, 23.6 at l=8, 29.3 at l=9, 31.0 at l=10, 35.0 at l=11, 39.4 at l=12, 39.2 at l=13, 38.7 at l=14, 39.9 at l=15, 41.6 at l=16, 43.6 at l=17, 41.9 at l=18, 41.5 at l=19, 39.8 at l=20, 39.7 at l=21, 38.9 at l=22, 35.9 at l=23, 30.9 at l=24, 28.0 at l=25, 24.8 at l=26, 22.1 at l=27, 21.0 at l=28, 20.4 at l=29, 20.1 at l=30, 20.0 at l=31, 18.5 at l=32. }
    \vspace{-1em}
    \caption{\textbf{Dependency Information per RemBERT Layer} via \depprobe{}'s structural, relational and parsing accuracy (UUAS, RelAcc, LAS) on EN-EWT (dev).}
    \label{fig:rembert-layers}
\end{figure}

\begin{figure}
    \centering
    \pdftooltip{\includegraphics[width=.47\textwidth]{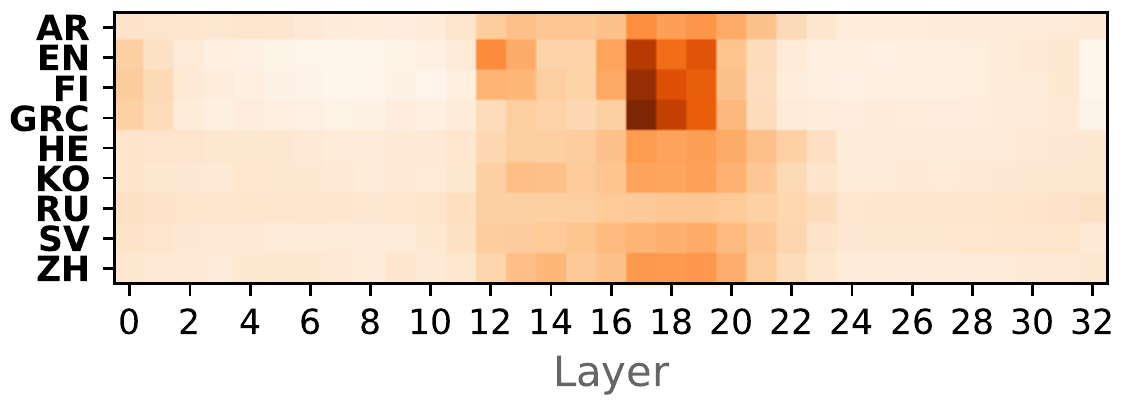}}{Screenreader Caption: Across all languages, the highest weighted layers are between 12 and 25, with substantially lower weights outside this range. A particularly high concentration can be found for EN, FI, GRC between layers 17 and 19.}
    \vspace{-.5em}
    \caption{\textbf{Per-language $\boldsymbol{\alpha}$ of RemBERT Layers} for \depprobe{} across all layer weights (dark $>$ light).}
    \label{fig:rembert-layer-weights}
\end{figure}

\paragraph{Layer-wise Probing} Prior probing studies have found dependency information to be concentrated around the middle layers of an LM \citep{hewitt-manning-2019-structural,tenney2018what,fayyaz-etal-2021-models}. Using EN-EWT \citep{silveira-etal-2014-gold}, we evaluate whether this holds for RemBERT's new architecture. Figure \ref{fig:rembert-layers} confirms that both dependency structural and relational information are most prominent around layer 17 of 32 as indicated by UUAS and relation classification accuracy (RelAcc) respectively. Combining the structural and relational information in \depprobe{} similarly leads to a peak of the LAS at the same layer while decreasing with further distance from the center.

Across all target languages, we next investigate whether probing a sum over the embeddings of all layers weighted by $\boldsymbol{\alpha} \in \mathbb{R}^{32}$ can boost extraction performance in RemBERT. The heavier weighting of middle layers by $\boldsymbol{\alpha}$, visible in Figure \ref{fig:rembert-layer-weights}, reaffirms a concentration of dependency information in the center. Contrasting probing work on prior models \citep{tenney2018what,kulmizev-etal-2020-neural}, using all layers does not increase the retrievable dependencies, with LAS differences $\pm$1 point. This further confirms that there is not a lack of dependency information in any specific layer, but that there is less within the encoder as a whole.

\begin{table}
\centering
\resizebox{.49\textwidth}{!}{
\begin{tabular}{lrrrrrrrrr}
\toprule
\textsc{Model} & AR & EN & FI & GRC & HE & KO & RU & SV & ZH \\
\midrule
\multirow{2}*{mBERT} & \textbf{65} & \textbf{74} & 65 & 46 & \textbf{69} & \textbf{58} & \textbf{68} & \textbf{65} & \textbf{58} \\
& \footnotesize{$\pm$.08} & \footnotesize{$\pm$.09} & \footnotesize{$\pm$.35} & \footnotesize{$\pm$.14} & \footnotesize{$\pm$.23} & \footnotesize{$\pm$.18} & \footnotesize{$\pm$.31} & \footnotesize{$\pm$.12} & \footnotesize{$\pm$.17} \\
\multirow{2}*{XLM-R} & 60 & 70 & \textbf{66} & 53 & 60 & 49 & 57 & 51 & 51 \\
 & \footnotesize{$\pm$.14} & \footnotesize{$\pm$.08} & \footnotesize{$\pm$.18} & \footnotesize{$\pm$.19} & \footnotesize{$\pm$.20} & \footnotesize{$\pm$.08} & \footnotesize{$\pm$.34} & \footnotesize{$\pm$.24} & \footnotesize{$\pm$.53} \\
\multirow{2}*{RemBERT} & 58 & 56 & 52 & \textbf{54} & 52 & 46 & 49 & 43 & 39 \\
& \footnotesize{$\pm$.12} & \footnotesize{$\pm$.22} & \footnotesize{$\pm$.15} & \footnotesize{$\pm$.18} & \footnotesize{$\pm$.05} & \footnotesize{$\pm$.14} & \footnotesize{$\pm$.04} & \footnotesize{$\pm$.08} & \footnotesize{$\pm$.24} \\
\bottomrule
\end{tabular}
}
\caption{\textbf{LAS of \bap{} Trained on Frozen LMs.} A biaffine attention parsing head is trained on top of frozen mBERT, XLM-R and RemBERT for each of the nine target languages ($\pm$ standard deviation).}
\label{tab:frozen-parsing}
\end{table}

\paragraph{Frozen Parsing} 
Our probing results show that linear subspaces in RemBERT contain less dependency information than prior LMs. However, \depprobe{}'s parametrization is kept intentionally simple and may therefore not be capturing non-linearly represented information that is useful during later fine-tuning. To evaluate this hypothesis, we train a full biaffine attention parsing head, but keep the underlying LM encoder frozen. This allows us to quantify the performance gains which come from inherent dependency information versus later task-specific fine-tuning.

Table \ref{tab:frozen-parsing} confirms our findings from \depprobe{} and shows that despite RemBERT outperforming mBERT and XLM-R when fully fine-tuned, it has substantially lower LAS across almost all languages when no full model fine-tuning is applied. This leads us to conclude that there indeed is less inherent dependency information in the newer model and that most performance gains must be occurring during task-specific full fine-tuning.

\paragraph{} Given that \depprobe{} extracts dependency structures reliably from LM architectures with different depths and embedding dimensionalities (e.g.\ RoBERTa\textsubscript{large} with 24 layers and $d$ = 1024 versus RuBERT\textsubscript{tiny} with 3 layers and $d$ = 312) as well as varying tokenization, optimization and pre-training data, the key difference in RemBERT appears to be embedding decoupling. The probe's linear formulation is not the limiting factor as the non-linear, biaffine attention head also produces less accurate parses when the LM's weights are frozen. Our analyses thus suggest that RemBERT's decoupled architecture contains less dependency information out-of-the-box, but follows prior patterns such as consolidating dependency information towards its middle layers and serving as strong initialization for parser training.

Lastly, RemBERT's larger number of tunable parameters compared to all other LM candidates may provide it further capacity, especially after full fine-tuning. As our probing methods are deliberately applied to the frozen representations of the encoder, it becomes especially important to consider the degree to which these embeddings may change after updating large parts of the model. Taking these limitations into account, the high correlations with respect to encoder ranking nonetheless enable a much more informed selection of LMs from a larger pool than was previously possible.

\section{Conclusion}\label{sec:conclusion}

To guide practitioners in their choice of LM encoder for the structured prediction task of dependency parsing, we leveraged a lightweight, linear \depprobe{} to quantify the latent syntactic information via the \textit{labeled} attachment score. Evaluating 46 pairs of multilingual/language-specific LMs and nine typologically diverse target treebanks, we found \depprobe{} to not only be efficient in its predictions, with orders of magnitude fewer trainable parameters, but to also be accurate 79--89\% of the time in predicting which LM will outperform another when used in a fully tuned parser. This allows for a substantially faster iteration over potential LM candidates, saving hours worth of compute in practice (Section \ref{sec:experiments}).

Our experiments further revealed surprising insights on the newly proposed RemBERT architecture: While particularly effective for multilingual dependency parsing when fully fine-tuned, it contains substantially less latent dependency information relative to prior widely-used models such as mBERT and XLM-R. Among its architectural differences, we identified embedding decoupling to be the most likely contributor, while added model capacity during fine-tuning may also improve final performance. Our analyses showed that despite containing less dependency information overall, RemBERT follows prior findings such as structure and syntactic relations being consolidated towards the middle layers. Given these consistencies, performance differences between decoupled LMs may be predictable using probes, but in absence of similar multilingual LMs using decoupled embeddings this effect remains to be studied (Section \ref{sec:rembert-analysis}).

Overall, the high efficiency and predictive power of ranking LM encoders via linear probing as well as the ease with which they can be analyzed---even when they encounter their limitations---offers immediate benefits to practitioners who have so far had to rely on their own intuitions when making a selection. This opens up avenues for future research by extending these methods to more tasks and LM architectures in order to enable better informed modeling decisions.

%
%
\section*{Acknowledgements}
We would like to thank the NLPnorth group for insightful discussions on this work, in particular Elisa Bassignana and Mike Zhang. Thanks also to ITU's High-performance Computing team. Finally, we thank the anonymous reviewers for their helpful feedback. This research is supported by the Independent Research Fund Denmark (Danmarks Frie Forskningsfond; DFF) grant number 9063-00077B.

\bibliography{anthology,references}
\bibliographystyle{acl_natbib}

%
%
\appendix

\section*{Appendices}\label{sec:appendix}

\section{Treebanks}\label{sec:treebank-details}

\begin{table}[h!]
\centering
\resizebox{.47\textwidth}{!}{
\begin{tabular}{lllr}
\toprule
\textsc{Target} & \textsc{Lang} & \textsc{Family} & \multicolumn{1}{l}{\textsc{Size}} \\
\midrule
AR-PADT & Arabic & Afro-Asiatic & 7.6k \\
EN-EWT & English & Indo-European & 16.6k \\
FI-TDT & Finnish & Uralic & 15.1k \\
GRC-PROIEL & Ancient Greek & Indo-European & 17.1k \\
HE-HTB & Hebrew & Afro-Asiatic & 6.2k \\
KO-GSD & Korean & Korean & 6.3k \\
RU-GSD & Russian & Indo-European & 5k \\
SV-Talbanken & Swedish & Indo-European & 6.0k \\
ZH-GSD & Chinese & Sino-Tibetan & 5.0k \\
\bottomrule
\end{tabular}
}
\caption{\label{tab:target-treebanks} \textbf{Target Treebanks} based on \citet{smith-etal-2018-investigation} with language family (\textsc{Family}) and total number of sentences (\textsc{Size}).}
\end{table}

Table \ref{tab:target-treebanks} lists the nine target treebanks based on the set by \citet{smith-etal-2018-investigation}: AR-PADT \citep{ar-padt}, EN-EWT \citep{silveira-etal-2014-gold},  FI-TDT \citep{pyysalo-etal-2015-universal}, GRC-PROIEL \citep{grc-proiel}, HE-HTB \citep{mcdonald-etal-2013-universal}, KO-GSD \citep{chun-etal-2018-building}, RU-GSD \citep{ru-gsd}, SV-Talbanken \citep{mcdonald-etal-2013-universal}, ZH-GSD \citep{zh-gsd}. We use these treebanks as provided in Universal Dependencies v2.9 \citep{ud29}. \depprobe{} and \bap{} are trained on each target's respective training split and are evaluated on the development split as this work aims to analyze general performance patterns instead of state-of-the-art performance.

\section{Experiment Setup}\label{sec:experiment-details}

\paragraph{\depprobe{}} is implemented in PyTorch v1.9.0 \citep{pytorch} and uses language models from the Transformers library v4.13.0 and the associated Model Hub \citep{wolf-etal-2020-transformers}. Following the structural probe by \citet{hewitt-manning-2019-structural}, each token which is split by the LM encoder into multiple subwords is mean-pooled. Similarly, we follow the original hyperparameter settings and set the structural subspace dimensionality to $b$ = 128 and use embeddings from the middle layer of each LM \citep{hewitt-manning-2019-structural,tenney2018what,fayyaz-etal-2021-models}. The structural loss is computed based on the absolute difference of the Euclidean distance between transformed word embeddings and the number of edges separating the words in the gold tree (see \citealp{hewitt-manning-2019-structural} for details). The relational loss is computed using cross entropy between the logits and gold head-child relation. Optimization uses AdamW \citep{loshchilov2018decoupled} with a learning rate of $10^{-3}$ which is reduced by a factor of 10 each time the loss plateaus. Early stopping is applied after three epochs without improvement and a maximum of 30 total epochs. With the only trainable parameters being the matrices $B$ and $L$, the model's footprint ranges between 51k and 190k parameters.

\paragraph{\bap{}} For the biaffine attention parser \citep{dozat2017} we use the implementation in the MaChAmp framework v0.3 \citep{van-der-goot-etal-2021-massive} with the default training schedule and hyperparameters. The number of trainable parameters depends on the LM encoder's size and ranges between 14M and 583M.

\paragraph{Analyses} For our analyses in Sections \ref{sec:experiments} and \ref{sec:rembert-analysis} we further make use of numpy v1.21.0 \citep{numpy}, SciPy v1.7.0 \citep{scipy} and Matplotlib v3.4.3 \citep{matplotlib}.

\paragraph{Training Details} Models are trained on an NVIDIA A100 GPU with 40GBs of VRAM and an AMD Epyc 7662 CPU. \bap{} requires around 1 h ($\pm$ 30 min). \depprobe{} can be trained in around 15 min ($\pm$ 5 min) with the embedding forward operation being most computationally expensive. The models use batches of size 32 and are initialized using the random seeds 692, 710 and 932.

\paragraph{Reproducibility} In order to ensure reproducibility and comparability with future work, we release our code and token-level predictions at \href{https://personads.me/x/naacl-2022-code}{https://personads.me/x/naacl-2022-code}.

\section{Detailed Results}\label{sec:result-details}

\begin{figure}[h!]
    \centering
    \begin{subfigure}[c]{.37\textwidth}
        \hspace{-.7em}
        \pdftooltip{\includegraphics[width=\textwidth]{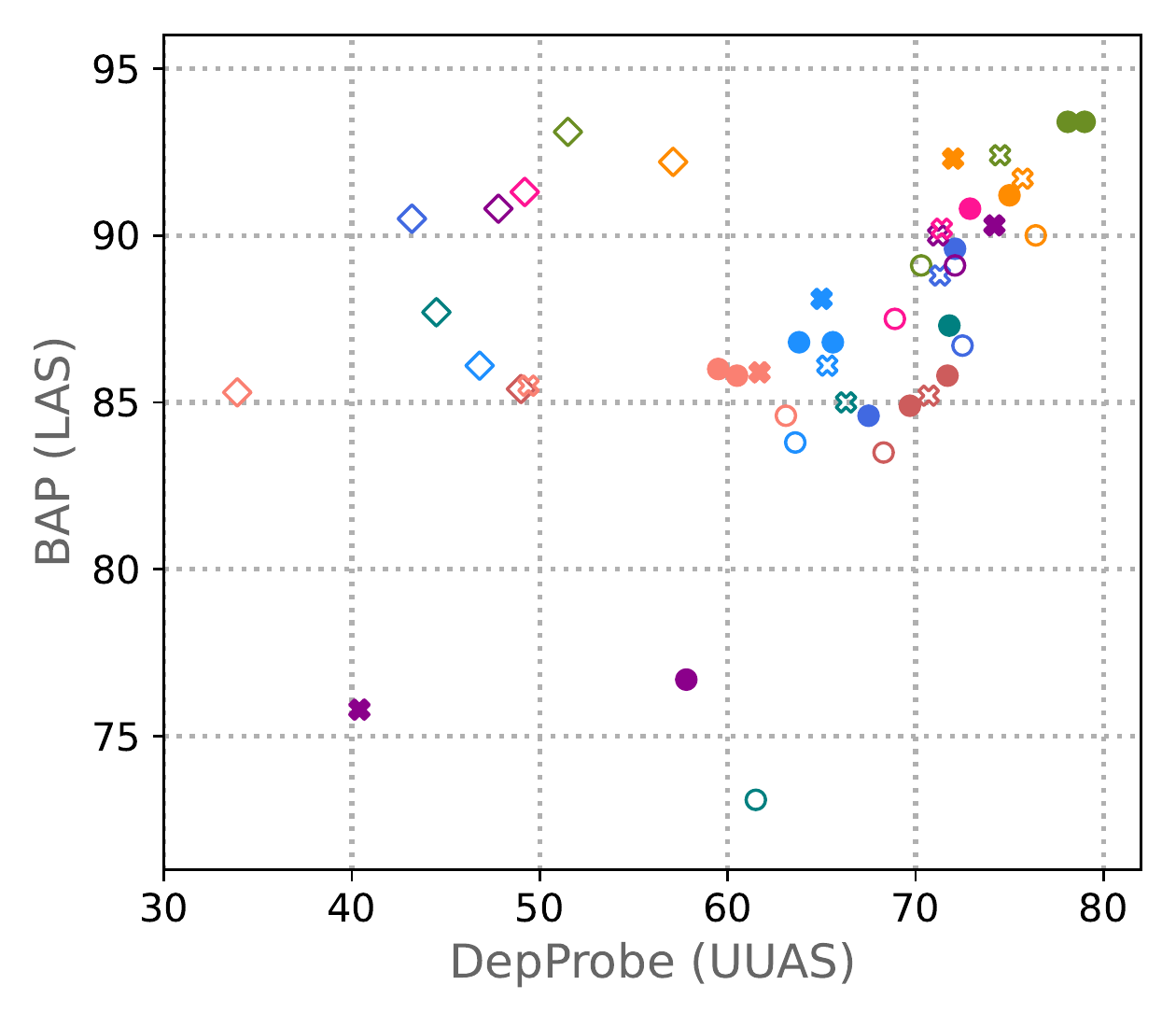}}{Screenreader Caption: BAP and DepProbe (UUAS). bert-base-multilingual-cased on AR-PADT: 83.50 and 68.30, xlm-roberta-base on AR-PADT: 85.20 and 70.70, google/rembert on AR-PADT: 85.40 and 49.00, aubmindlab/bert-base-arabertv02 on AR-PADT: 85.80 and 71.70, asafaya/bert-base-arabic on AR-PADT: 84.90 and 69.70, bert-base-multilingual-cased on EN-EWT: 90.00 and 76.40, xlm-roberta-base on EN-EWT: 91.70 and 75.70, google/rembert on EN-EWT: 92.20 and 57.10, bert-base-uncased on EN-EWT: 91.20 and 75.00, roberta-large on EN-EWT: 92.30 and 72.00, bert-base-multilingual-cased on FI-TDT: 89.10 and 70.30, xlm-roberta-base on FI-TDT: 92.40 and 74.50, google/rembert on FI-TDT: 93.10 and 51.50, TurkuNLP/bert-base-finnish-uncased-v1 on FI-TDT: 93.40 and 79.00, TurkuNLP/bert-base-finnish-cased-v1 on FI-TDT: 93.40 and 78.10, bert-base-multilingual-cased on GRC-PROIEL: 73.10 and 61.50, xlm-roberta-base on GRC-PROIEL: 85.00 and 66.30, google/rembert on GRC-PROIEL: 87.70 and 44.50, pranaydeeps/Ancient-Greek-BERT on GRC-PROIEL: 87.30 and 71.80, nlpaueb/bert-base-greek-uncased-v1 on HE-HTB: 84.60 and 67.50, bert-base-multilingual-cased on HE-HTB: 86.70 and 72.50, xlm-roberta-base on HE-HTB: 88.80 and 71.30, google/rembert on HE-HTB: 90.50 and 43.20, onlplab/alephbert-base on HE-HTB: 89.60 and 72.10, bert-base-multilingual-cased on KO-GSD: 83.80 and 63.60, xlm-roberta-base on KO-GSD: 86.10 and 65.30, google/rembert on KO-GSD: 86.10 and 46.80, klue/bert-base on KO-GSD: 86.80 and 65.60, klue/roberta-large on KO-GSD: 88.10 and 65.00, kykim/bert-kor-base on KO-GSD: 86.80 and 63.80, bert-base-multilingual-cased on RU-GSD: 89.10 and 72.10, xlm-roberta-base on RU-GSD: 90.00 and 71.20, google/rembert on RU-GSD: 90.80 and 47.80, cointegrated/rubert-tiny on RU-GSD: 76.70 and 57.80, sberbank-ai/ruRoberta-large on RU-GSD: 90.30 and 74.20, blinoff/roberta-base-russian-v0 on RU-GSD: 75.80 and 40.40, bert-base-multilingual-cased on SV-Talbanken: 87.50 and 68.90, xlm-roberta-base on SV-Talbanken: 90.20 and 71.40, google/rembert on SV-Talbanken: 91.30 and 49.20, KB/bert-base-swedish-cased on SV-Talbanken: 90.80 and 72.90, bert-base-multilingual-cased on ZH-GSD: 84.60 and 63.10, xlm-roberta-base on ZH-GSD: 85.50 and 49.40, google/rembert on ZH-GSD: 85.30 and 33.90, bert-base-chinese on ZH-GSD: 85.80 and 60.50, hfl/chinese-bert-wwm-ext on ZH-GSD: 86.00 and 59.50, hfl/chinese-roberta-wwm-ext on ZH-GSD: 85.90 and 61.70.}
    \end{subfigure}
    \hspace{-1em}
    \begin{subfigure}[c]{.08\textwidth}
        \includegraphics[width=\textwidth]{figures/legend-models.pdf}\\[.8em]
        \includegraphics[width=.7\textwidth]{figures/legend-languages.pdf}
    \end{subfigure}
    \caption{\textbf{UUAS of \depprobe{} in relation to \bap{}} across nine language targets (dev) using language-specific and multilingual LM encoders of different architecture types.}
    \label{fig:scatter-uuas-results}
\end{figure}

Tables \ref{tab:results-ar}--\ref{tab:results-zh} list exact LAS and standard deviations for each experiment in Section \ref{sec:experiments}'s Figure \ref{fig:scatter-results} in addition to the HuggingFace Model Hub IDs of the LMs used in each of the 46 setups as well as their number of layers, embedding dimensionality $d$ and total number of parameters. In addition, Figure \ref{fig:scatter-uuas-results} shows UUAS for all setups, equivalent to only probing structurally \citep{hewitt-manning-2019-structural} for unlabeled, undirected dependency trees.

%
%
\begin{table*}
\centering
\resizebox{.95\textwidth}{!}{
\begin{tabular}{llrrrrr}
\toprule
\textsc{Models} & \textsc{Source} & \multicolumn{1}{c}{\textsc{Layers}} & \multicolumn{1}{c}{\textsc{Emb $d$}} & \multicolumn{1}{c}{\textsc{Params}} & \multicolumn{1}{l}{\bap{}} & \multicolumn{1}{l}{\depprobe{}} \\
\midrule
\texttt{bert-base-multilingual-cased} & \citet{devlin-etal-2019-bert} & 12 & 768 & 178M & 83.5\footnotesize{$\pm$0.2} & 54.8\footnotesize{$\pm$0.6} \\
\texttt{xlm-roberta-base} & \citet{conneau-etal-2020-unsupervised} & 12 & 768 & 278M & 85.2\footnotesize{$\pm$0.1} & 57.2\footnotesize{$\pm$0.1} \\
\texttt{google/rembert} & \citet{chung2021rethinking} & 32 & 1152 & 576M & 85.4\footnotesize{$\pm$0.2} & 20.7\footnotesize{$\pm$0.1} \\
\midrule
\texttt{aubmindlab/bert-base-arabertv02} & \citet{antoun-etal-2020-arabert} & 12 & 768 & 135M & 85.8\footnotesize{$\pm$0.1} & 59.0\footnotesize{$\pm$0.1} \\
\texttt{asafaya/bert-base-arabic} & \citet{safaya-etal-2020-kuisail} & 12 & 768 & 111M & 84.9\footnotesize{$\pm$0.1} & 57.0\footnotesize{$\pm$0.2} \\
\bottomrule
\end{tabular}
}
\caption{\label{tab:results-ar} \textbf{LAS on AR-PADT (Dev)} using \bap{} and \depprobe{} with different LMs ($\pm$ standard deviation).}
\end{table*}

%
%
\begin{table*}
\centering
\resizebox{.95\textwidth}{!}{
\begin{tabular}{llrrrrr}
\toprule
\textsc{Models} & \textsc{Source} & \multicolumn{1}{c}{\textsc{Layers}} & \multicolumn{1}{c}{\textsc{Emb $d$}} & \multicolumn{1}{c}{\textsc{Params}} & \multicolumn{1}{l}{\bap{}} & \multicolumn{1}{l}{\depprobe{}} \\
\midrule
\texttt{bert-base-multilingual-cased} & \citet{devlin-etal-2019-bert} & 12 & 768 & 178M & 90.0\footnotesize{$\pm$0.1} & 64.5\footnotesize{$\pm$0.3} \\
\texttt{xlm-roberta-base} & \citet{conneau-etal-2020-unsupervised} & 12 & 768 & 278M & 91.7\footnotesize{$\pm$0.2} & 64.8\footnotesize{$\pm$0.1} \\
\texttt{google/rembert} & \citet{chung2021rethinking} & 32 & 1152 & 576M & 92.2\footnotesize{$\pm$0.0} & 41.6\footnotesize{$\pm$0.3} \\
\midrule
\texttt{bert-base-uncased} & \citet{devlin-etal-2019-bert} & 12 & 768 & 109M & 91.2\footnotesize{$\pm$0.1} & 63.4\footnotesize{$\pm$0.3} \\
\texttt{roberta-large} & \citet{liu2019roberta} & 24 & 1024 & 355M & 92.3\footnotesize{$\pm$0.2} & 59.9\footnotesize{$\pm$0.2} \\
\bottomrule
\end{tabular}
}
\caption{\label{tab:results-en} \textbf{LAS on EN-EWT (Dev)} using \bap{} and \depprobe{} with different LMs ($\pm$ standard deviation).}
\end{table*}

%
%
\begin{table*}
\centering
\resizebox{.95\textwidth}{!}{
\begin{tabular}{llrrrrr}
\toprule
\textsc{Models} & \textsc{Source} & \multicolumn{1}{c}{\textsc{Layers}} & \multicolumn{1}{c}{\textsc{Emb $d$}} & \multicolumn{1}{c}{\textsc{Params}} & \multicolumn{1}{l}{\bap{}} & \multicolumn{1}{l}{\depprobe{}} \\
\midrule
\texttt{bert-base-multilingual-cased} & \citet{devlin-etal-2019-bert} & 12 & 768 & 178M & 89.1\footnotesize{$\pm$0.2} & 54.5\footnotesize{$\pm$0.4} \\
\texttt{xlm-roberta-base} & \citet{conneau-etal-2020-unsupervised} & 12 & 768 & 278M & 92.4\footnotesize{$\pm$0.1} & 62.4\footnotesize{$\pm$0.2} \\
\texttt{google/rembert} & \citet{chung2021rethinking} & 32 & 1152 & 576M & 93.1\footnotesize{$\pm$0.1} & 30.8\footnotesize{$\pm$0.1} \\
\midrule
\texttt{TurkuNLP/bert-base-finnish-uncased-v1} & \citet{virtanen2019finbert} & 12 & 768 & 125M & 93.4\footnotesize{$\pm$0.1} & 68.9\footnotesize{$\pm$0.3} \\
\texttt{TurkuNLP/bert-base-finnish-cased-v1} & \citet{virtanen2019finbert} & 12 & 768 & 125M & 93.4\footnotesize{$\pm$0.1} & 67.5\footnotesize{$\pm$0.4} \\
\bottomrule
\end{tabular}
}
\caption{\label{tab:results-fi} \textbf{LAS on FI-TDT (Dev)} using \bap{} and \depprobe{} with different LMs ($\pm$ standard deviation).}
\end{table*}

%
%
\begin{table*}
\centering
\resizebox{.95\textwidth}{!}{
\begin{tabular}{llrrrrr}
\toprule
\textsc{Models} & \textsc{Source} & \multicolumn{1}{c}{\textsc{Layers}} & \multicolumn{1}{c}{\textsc{Emb $d$}} & \multicolumn{1}{c}{\textsc{Params}} & \multicolumn{1}{l}{\bap{}} & \multicolumn{1}{l}{\depprobe{}} \\
\midrule
\texttt{bert-base-multilingual-cased} & \citet{devlin-etal-2019-bert} & 12 & 768 & 178M & 73.1\footnotesize{$\pm$0.1} & 41.6\footnotesize{$\pm$0.5} \\
\texttt{xlm-roberta-base} & \citet{conneau-etal-2020-unsupervised} & 12 & 768 & 278M & 85.0\footnotesize{$\pm$0.2} & 51.1\footnotesize{$\pm$0.2} \\
\texttt{google/rembert} & \citet{chung2021rethinking} & 32 & 1152 & 576M & 87.7\footnotesize{$\pm$0.1} & 15.3\footnotesize{$\pm$0.1} \\
\midrule
\texttt{pranaydeeps/Ancient-Greek-BERT} & \citet{singh-etal-2021-pilot} & 12 & 768 & 113M & 87.3\footnotesize{$\pm$0.1} & 60.0\footnotesize{$\pm$0.0} \\
\texttt{nlpaueb/bert-base-greek-uncased-v1} & \citet{koutsikakis2020greekbert} & 12 & 768 & 113M & 84.6\footnotesize{$\pm$0.3} & 53.9\footnotesize{$\pm$0.1} \\
\bottomrule
\end{tabular}
}
\caption{\label{tab:results-grc} \textbf{LAS on GRC-PROIEL (Dev)} using \bap{} and \depprobe{} with different LMs ($\pm$ standard deviation).}
\end{table*}

%
%
\begin{table*}
\centering
\resizebox{.95\textwidth}{!}{
\begin{tabular}{llrrrrr}
\toprule
\textsc{Models} & \textsc{Source} & \multicolumn{1}{c}{\textsc{Layers}} & \multicolumn{1}{c}{\textsc{Emb $d$}} & \multicolumn{1}{c}{\textsc{Params}} & \multicolumn{1}{l}{\bap{}} & \multicolumn{1}{l}{\depprobe{}} \\
\midrule
\texttt{bert-base-multilingual-cased} & \citet{devlin-etal-2019-bert} & 12 & 768 & 178M & 86.7\footnotesize{$\pm$0.2} & 60.2\footnotesize{$\pm$0.6} \\
\texttt{xlm-roberta-base} & \citet{conneau-etal-2020-unsupervised} & 12 & 768 & 278M & 88.8\footnotesize{$\pm$0.1} & 59.2\footnotesize{$\pm$0.3} \\
\texttt{google/rembert} & \citet{chung2021rethinking} & 32 & 1152 & 576M & 90.5\footnotesize{$\pm$0.1} & 11.6\footnotesize{$\pm$0.4} \\
\midrule
\texttt{onlplab/alephbert-base} & \citet{seker2021alephbert} & 12 & 768 & 126M & 89.6\footnotesize{$\pm$0.1} & 61.4\footnotesize{$\pm$0.2} \\
\bottomrule
\end{tabular}
}
\caption{\label{tab:results-he} \textbf{LAS on HE-HTB (Dev)} using \bap{} and \depprobe{} with different LMs ($\pm$ standard deviation).}
\end{table*}

%
%
\begin{table*}
\centering
\resizebox{.95\textwidth}{!}{
\begin{tabular}{llrrrrr}
\toprule
\textsc{Models} & \textsc{Source} & \multicolumn{1}{c}{\textsc{Layers}} & \multicolumn{1}{c}{\textsc{Emb $d$}} & \multicolumn{1}{c}{\textsc{Params}} & \multicolumn{1}{l}{\bap{}} & \multicolumn{1}{l}{\depprobe{}} \\
\midrule
\texttt{bert-base-multilingual-cased} & \citet{devlin-etal-2019-bert} & 12 & 768 & 178M & 83.8\footnotesize{$\pm$0.2} & 46.6\footnotesize{$\pm$0.2} \\
\texttt{xlm-roberta-base} & \citet{conneau-etal-2020-unsupervised} & 12 & 768 & 278M & 86.1\footnotesize{$\pm$0.1} & 49.4\footnotesize{$\pm$0.3} \\
\texttt{google/rembert} & \citet{chung2021rethinking} & 32 & 1152 & 576M & 86.1\footnotesize{$\pm$0.2} & 15.9\footnotesize{$\pm$0.3} \\
\midrule
\texttt{klue/bert-base} & \citet{park2021klue} & 12 & 768 & 111M & 86.8\footnotesize{$\pm$0.0} & 51.0\footnotesize{$\pm$0.1} \\
\texttt{klue/roberta-large} & \citet{park2021klue} & 24 & 1024 & 337M & 88.1\footnotesize{$\pm$0.3} & 48.8\footnotesize{$\pm$0.5} \\
\texttt{kykim/bert-kor-base} & \citet{kim2020lmkor} & 12 & 768 & 118M & 86.8\footnotesize{$\pm$0.1} & 46.9\footnotesize{$\pm$0.4} \\
\bottomrule
\end{tabular}
}
\caption{\label{tab:results-ko} \textbf{LAS on KO-GSD (Dev)} using \bap{} and \depprobe{} with different LMs ($\pm$ standard deviation).}
\end{table*}

%
%
\begin{table*}
\centering
\resizebox{.95\textwidth}{!}{
\begin{tabular}{llrrrrr}
\toprule
\textsc{Models} & \textsc{Source} & \multicolumn{1}{c}{\textsc{Layers}} & \multicolumn{1}{c}{\textsc{Emb $d$}} & \multicolumn{1}{c}{\textsc{Params}} & \multicolumn{1}{l}{\bap{}} & \multicolumn{1}{l}{\depprobe{}} \\
\midrule
\texttt{bert-base-multilingual-cased} & \citet{devlin-etal-2019-bert} & 12 & 768 & 178M & 89.1\footnotesize{$\pm$0.1} & 60.7\footnotesize{$\pm$0.1} \\
\texttt{xlm-roberta-base} & \citet{conneau-etal-2020-unsupervised} & 12 & 768 & 278M & 90.0\footnotesize{$\pm$0.2} & 59.9\footnotesize{$\pm$1.1} \\
\texttt{google/rembert} & \citet{chung2021rethinking} & 32 & 1152 & 576M & 90.8\footnotesize{$\pm$0.0} & 26.0\footnotesize{$\pm$0.2} \\
\midrule
\texttt{cointegrated/rubert-tiny} & \citet{dale2021rubert} & 3 & 312 & 11M & 76.7\footnotesize{$\pm$0.1} & 41.5\footnotesize{$\pm$0.6} \\
\texttt{sberbank-ai/ruRoberta-large} & \citet{sberdevices2021rubert} & 24 & 1024 & 355M & 90.3\footnotesize{$\pm$0.3} & 63.2\footnotesize{$\pm$0.4} \\
\texttt{blinoff/roberta-base-russian-v0} & \citet{blinov2021robertaru} & 12 & 768 & 124M & 75.8\footnotesize{$\pm$0.0} & 15.6\footnotesize{$\pm$0.2} \\
\bottomrule
\end{tabular}
}
\caption{\label{tab:results-ru} \textbf{LAS on RU-GSD (Dev)} using \bap{} and \depprobe{} with different LMs ($\pm$ standard deviation).}
\end{table*}

%
%
\begin{table*}
\centering
\resizebox{.95\textwidth}{!}{
\begin{tabular}{llrrrrr}
\toprule
\textsc{Models} & \textsc{Source} & \multicolumn{1}{c}{\textsc{Layers}} & \multicolumn{1}{c}{\textsc{Emb $d$}} & \multicolumn{1}{c}{\textsc{Params}} & \multicolumn{1}{l}{\bap{}} & \multicolumn{1}{l}{\depprobe{}} \\
\midrule
\texttt{bert-base-multilingual-cased} & \citet{devlin-etal-2019-bert} & 12 & 768 & 178M & 87.5\footnotesize{$\pm$0.1} & 55.5\footnotesize{$\pm$0.2} \\
\texttt{xlm-roberta-base} & \citet{conneau-etal-2020-unsupervised} & 12 & 768 & 278M & 90.2\footnotesize{$\pm$0.1} & 59.1\footnotesize{$\pm$0.2} \\
\texttt{google/rembert} & \citet{chung2021rethinking} & 32 & 1152 & 576M & 91.3\footnotesize{$\pm$0.3} & 31.7\footnotesize{$\pm$0.3} \\
\midrule
\texttt{KB/bert-base-swedish-cased} & \citet{malmsten2020swedishbert} & 12 & 768 & 125M & 90.8\footnotesize{$\pm$0.1} & 61.7\footnotesize{$\pm$0.2} \\
\bottomrule
\end{tabular}
}
\caption{\label{tab:results-sv} \textbf{LAS on SV-Talbanken (Dev)} using \bap{} and \depprobe{} with different LMs ($\pm$ standard deviation).}
\end{table*}

%
%
\begin{table*}
\centering
\resizebox{.95\textwidth}{!}{
\begin{tabular}{llrrrrr}
\toprule
\textsc{Models} & \textsc{Source} & \multicolumn{1}{c}{\textsc{Layers}} & \multicolumn{1}{c}{\textsc{Emb $d$}} & \multicolumn{1}{c}{\textsc{Params}} & \multicolumn{1}{l}{\bap{}} & \multicolumn{1}{l}{\depprobe{}} \\
\midrule
\texttt{bert-base-multilingual-cased} & \citet{devlin-etal-2019-bert} & 12 & 768 & 178M & 84.6\footnotesize{$\pm$0.4} & 49.1\footnotesize{$\pm$0.4} \\
\texttt{xlm-roberta-base} & \citet{conneau-etal-2020-unsupervised} & 12 & 768 & 278M & 85.5\footnotesize{$\pm$0.3} & 30.3\footnotesize{$\pm$0.1} \\
\texttt{google/rembert} & \citet{chung2021rethinking} & 32 & 1152 & 576M & 85.3\footnotesize{$\pm$0.2} & 5.2\footnotesize{$\pm$0.1} \\
\midrule
\texttt{bert-base-chinese} & \citet{devlin-etal-2019-bert} & 12 & 768 & 102M & 85.8\footnotesize{$\pm$0.1} & 46.4\footnotesize{$\pm$0.1} \\
\texttt{hfl/chinese-bert-wwm-ext} & \citet{cui2021zhbert} & 12 & 768 & 102M & 86.0\footnotesize{$\pm$0.3} & 45.8\footnotesize{$\pm$0.3} \\
\texttt{hfl/chinese-roberta-wwm-ext} & \citet{cui2021zhbert} & 12 & 768 & 102M & 85.9\footnotesize{$\pm$0.3} & 47.7\footnotesize{$\pm$0.4} \\
\bottomrule
\end{tabular}
}
\caption{\label{tab:results-zh} \textbf{LAS on ZH-GSD (Dev)} using \bap{} and \depprobe{} with different LMs ($\pm$ standard deviation).}
\end{table*}

\end{document}